\documentclass[runningheads]{llncs}
\usepackage{cite}
\usepackage[dvips]{graphicx}
\graphicspath{{eps/}}
\usepackage{upgreek}
\DeclareGraphicsExtensions{.eps}
\usepackage[cmex10]{amsmath}
\usepackage{amsfonts}
\usepackage{amssymb}
\usepackage{url}
\newcommand{\keywords}[1]{\par\addvspace\baselineskip
\noindent\keywordname\enspace\ignorespaces#1}
\usepackage{multirow}
\usepackage{subcaption}
\usepackage{algorithm2e}
\usepackage{verbatim}
\usepackage{fixltx2e}
\usepackage{color}

\begin{document}

\title{{Partitioned Shape Modeling with \emph{On-the-Fly} Sparse Appearance Learning for {{Anterior Visual Pathway Segmentation}}}}
\titlerunning{Partitioned Joint Shape Modeling with Sparse Appearance Learning}
\author{Awais Mansoor, Juan J. Cerrolaza, Robert A. Avery, Marius G. Linguraru}
\authorrunning{Mansoor et al.}
\institute{Children's National Medical Center, \\111 Michigan Avenue NW, Washington, DC 20010, USA.\\
\email{awais.mansoor@gmail.com}}
\maketitle

\begin{abstract}
{{MRI quantification of cranial nerves such as anterior visual pathway (AVP) in MRI is challenging due to their thin small size, structural variation along its path, and adjacent anatomic structures}}. Segmentation of pathologically abnormal optic nerve (e.g. optic nerve glioma) poses additional challenges due to changes in its shape at unpredictable locations. In this work, we propose a partitioned joint statistical shape model approach with sparse appearance learning for the segmentation of healthy and pathological AVP. Our main contributions are: (1) optimally partitioned statistical shape models for the AVP based on regional shape variations for greater local flexibility of statistical shape model; {{(2) refinement model to accommodate pathological regions as well as areas of subtle variation by training the model \emph{on-the-fly} using the initial segmentation obtained in (1);}} (3) hierarchical deformable framework to incorporate scale information in partitioned shape and appearance models. Our method, entitled PAScAL (PArtitioned Shape and Appearance Learning), was evaluated on 21 MRI scans (15 healthy + 6 glioma cases) from pediatric patients (ages 2-17). {{The experimental results show that the proposed localized shape and sparse appearance-based learning approach significantly outperforms segmentation approaches in the analysis of pathological data}}. 
\end{abstract}

\section{Introduction}
MRI is a widely used non-invasive technique for studying and characterizing diseases of the optic pathway such as optic neuritis, multiple sclerosis, and optic pathway glioma (OPG) \cite{chan2007optic}. OPGs are low grade astrocytomas inherent to the AVP (i.e., optic nerve, chiasm and tracts). OPGs occur in $20\%$ of children with neurofibromatosis type 1 (NF1), a very common genetic disorder {that carries increased risk of tumors in the nervous system}. The disease course is variable, as these tumors may demonstrate several distinct periods of growth, stability or regression. {{Currently, no quantitative imaging criteria exist to define OPGs secondary to NF1}}. Non-invasive computer-aided quantification of these changes can not only eliminate excessive physician’s effort to segment these regions but also increases the precision of volume measures. However, automatic segmentation of cranial nerve pathways including AVP from MRI is challenging due to their thin-long shape and varying appearances. A few non-invasive automated methods to segment AVP from radiological images have been reported in the literature previously with modest success. Bekes et al. \cite{bekes2008735} proposed a geometrical model based approach; however, their approach's reproducibility is found to be less than $50\%$. Noble et al. \cite{noble2011877} presented a hybrid approach using a deformable model with level set method to segment the optic nerves and the chiasm; {{however, the method was tested only on healthy cases}. Recently, Yang et al. \cite{Xue2014109} developed a partitioned approach to healthy AVP segmentation by dividing the pathway into various shape homogenous segments and modeling each segment independently. The local appearance information in their approach was encoded using the normalized derivatives, three class fuzzy c-means, and spherical flux}. {{The approach was the first attempt to accommodate local shape and appearance variation for healthy AVP segmentation}}; the method, although promising, did not provide any objective criteria on the optimal number of partitions. Moreover, the approach did not accommodate local appearance characteristics along the nerve boundary that are particularly important in pathological cases. 

\begin{figure}[!t]
\centering
\includegraphics[width=0.9\linewidth]{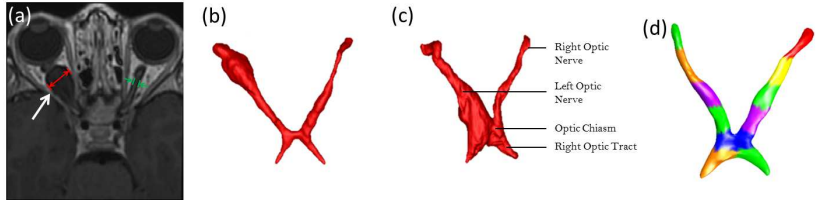}
\caption{\small{(a) MRI scan with a healthy (\emph{left}) and a gliomic (\emph{right}) optic nerve. The maximum diameter of OPG nerve is $9.54$ mm and $1.15$ mm for the healthy nerve of the same patient. (b)-(c) renderings of typical OPG cases in the optic nerve. {{(b) shows OPG in the distal region of left optic nerve, (c) shows one in the proximal region. (d) Shape consistent partitioning of a healthy AVP produced by PAScAL.}}}}
\label{fig:introduction}
\vspace{-.2in}
\end{figure}

Depending on severity, pathological AVPs can have a drastically different local shape and appearance characteristics than healthy ones, thus failing the shape model based segmentation methods in cranial nerve pathways. To illustrate, Fig \ref{fig:introduction}(a) demonstrates a healthy optic nerve along with a contralateral optic nerve having OPG. Fig \ref{fig:introduction} (b)-(c) show the renderings of cases with OPG in optic nerve region. In this paper, we propose, \emph{PAScAL}, an optimally partitioned statistical shape model with sparse appearance learning for the segmentation of AVPs for both healthy and pathological cases. The challenge of segmenting larger anatomical structures with pathologies have been addressed numerously in the literature \cite{mansoor2014generic}. However, development of similar approaches for {{smaller vascular structures, such as the AVP,}} have traditionally been ignored. {{By illustrating the robustness of PAScAL to segment AVP with OPG, we demonstrate the applicability of the proposed method in segmenting other anatomical structures of similar characteristics.}}
   
\section{Methods}
\begin{figure}[!t]
\centering
\includegraphics[width=0.9\linewidth]{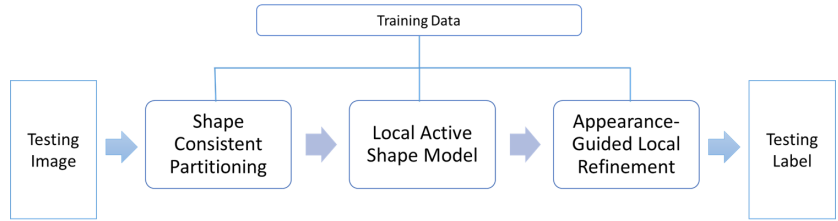}
\caption{\small{Flow diagram of the PAScAL approach to optic nerve segmentation.}}
\label{fig:flowdiagram}
\end{figure}

We propose a hierarchical joint partitioned shape model and sparse appearance learning to automatically segment the AVP from MRI scans of the head. During \textbf{the training stage} automatically selected landmarks from healthy cases are first clustered into various shape-consistent overlapping partitions thus creating individual simplistic shape and appearance models for each partition. The individually learned models are used to produce the initial segmentation of AVP using the partitioned active shape model (ASM\textsubscript{p}) described in Section \ref{sec:asmp}. In \textbf{the testing stage}, the learned ASM\textsubscript{p} is iteratively fitted to new data using the appearance guided model. A refinement stage follows to accommodate local appearance features particularly important in cases with pathologies (e.g. OPG): a sparse local appearance dictionary is learned \emph{on-the-fly} from the testing image for each partition using the initial segmentation as training data acquired from the test image in real-time. Through these steps, PAScAL is adapting to each testing set to compensate for the difficulties with off-line training for pathological cases due to the unpredictable location, shape, and appearance of OPG. PAScAL is summarized in Fig. \ref{fig:flowdiagram}. Details of the proposed method are provided in the subsequent sections.

\subsection{Shape consistent agglomerative hierarchical landmark partitioning}
In the beginning, the annotated landmarks are grouped by using a modification of the agglomerative hierarchical clustering method proposed by Cerrolaza et al. \cite{cerrolaza2015automatic}, minimizing the following objective function:
\begin{equation}
J\left( \Omega  \right) = \underbrace {\alpha \int\limits_\Omega  {{{\left( {\frac{{\left| {{\mathbf{V}_\Omega } \times {\mathbf{V}_l}} \right|}}{{\left| {{\mathbf{V}_l}} \right|}}} \right)}^2}\frac{{{L_{\max }}}}{{\left| {{\mathbf{V}_l}} \right|}}dl} }_{{\text{Colinearity term}}} + \underbrace {\left( {1 - \alpha } \right)\left( {1 - \frac{{\int\limits_\Omega  {dl} }}{{\int\limits_\mathbf{S} {dl}}}} \right)}_{{\text{Maximum area constraint}}}
\end{equation}
where $\mathbf{S}$ is the set of all landmarks over the AVP and $\Omega\subset\mathbf{S}$ denotes the local shape to be sub-partitioned into optimal set of clusters. $\mathbf{V}_\Omega$ denotes the dominant direction in $\Omega$. $\mathbf{V}_l$ is the deformation vector for landmark $l$ obtained through well known point distribution model by Cootes et. al \cite{cootes2001236} over $\mathbf{S}$. $\alpha\in[0,1]$ is the coefficient that controls the relative weights ($\alpha$ is set to $0.8$ in our experiments) and ${L_{\max }} = \mathop {\max }\limits_\mathbf{S} \left\{ {\left\| {{V_l}} \right\|} \right\}$. We define the optimal number of partitions based on shape similarities calculated using a tailored Silhouette coefficient score. Specifically, let $\Omega_p$ denotes the set of landmarks for the shape partition $p$ containing the landmark $l$ and $\Omega_{p-l}$ denotes the set of landmarks for the same shape $p$ with landmark excluded then the contribution of the landmark $l$ in partition $p$ is defined as $a_{p,l}=J(\Omega_{p})-J(\Omega_{p-l}) \in \{0,1\}$. A large $a_{p,l}$ denotes higher dissimilarity between the landmark $l$ and the shape $\Omega_p$. The cost of including landmark $l$ to a partition $p$ is similarly defined as $b_{p,l}=J(\Omega_{p+l})-J(\Omega_{p})$. Then the optimal number of partitions $p_\text{opt}$ are found by maximizing: 
$
\begin{aligned}
 & \underset{\Omega}{\textit{maximize}}
 & \frac{1}{|l|}\sum_{p=1}^{|l|}\frac{f_p(b_l)-f_p(a_l)}{\text{max}(f_p(a_l),f_p(b_l))} \\
 \end{aligned}
$, where $f(.)$ is the logistic sigmoid function, $|l|$ is the total number of landmarks. To ensure that adjacent partitions are connected, an overlapping region is introduced by sharing the boundary landmarks of these partitions. During the shape model fitting, the shape parameters of the overlapping landmarks are calculated using the parameters of the overlapping shapes. {{Fig. \ref{fig:introduction}(d) demonstrates the proposed agglomerative hierarchical landmark partitioning approach.}}

\subsection{Landmark weighted partitioned active shape model fitting}
\label{sec:asmp}
Once the shape partitions are generated, ASM\textsubscript{p} is performed on the individual shapes in the partitioned hyperspace. In order to adapt to local appearance characteristics, following set of appearance features are used to create overlapping partitioned statistical appearance models for each partition: (i) the intensities of neighboring voxels of each landmark, (ii) the three-class fuzzy c-means filter to robustly delineate both tissues in dark as well as bright foregrounds (as explained before, the AVP passes through neighboring tissues of varying contrasts), and (iii) spherical flux to exploit the \emph{vessel-like} characteristics. AVP has varying contrast in different regions (i.e, fatty regions has better contrast appearance with optic nerve than gray matter) thus we assigned different levels of confidence for the reliability of landmarks. Specifically, for each landmark in the training set, the covariance $\Sigma$ of these features is calculated across the training examples under the assumption that the lower the variance of the appearance profile of a landmark, the higher would be our confidence in the landmark. The weight $w_l$ of a landmark $l$ can therefore be calculated as: ${w_l} = \frac{1}{{\left( {1 + tr\left( {{\Sigma_l}} \right)} \right)}}$, where $tr()$ denotes the trace of a matrix. The shape parameters for a partition $p$ can be computed as ${b_p} = {\left( {\varphi _p^T{W_p}\varphi _p^T} \right)^{ - 1}}\varphi _p^T{W_p}\left( {{x_p} - \overline {{x_p}} } \right)$, where $\varphi_p$ is the eigenvector matrix, $x_p$ is the aligned training shape vector, $\overline {{x_p}}$ is the mean shape vector, and $W_p$ is the diagonal weight matrix of landmarks belonging partition $p$. 

\subsection{\emph{On-the-fly} sparse appearance learning}
\label{sec:fly}
Pathologies can result in changes in shape and appearance of AVP at unpredictable locations (Fig. \ref{fig:introduction}). {{Statistical shape models have been very successful in segmenting healthy organs; however, they struggle to accommodate cases where the shape of the target structure cannot be predicted through training, such as in the cases of OPG. Feature-based approaches have demonstrated superior performance in segmentation of pathological tissues \cite{mansoor2014generic}; however, off-line feature-based training of pathological cases mostly fails due to large variations, in both shape and appearance, for pathological cases}}. To address these challenges, we present a novel \emph{on-the-fly} learning approach by using the initial delineation of the test image obtained in the previous section as training to learn an appearance dictionary in real-time. Specifically, let $R_v(p)$ be a $m\times m\times k$ image patch extracted from within the initial partition $p$ centered at voxel $v\in\mathbb{R}^3$. Equal number of patches are extracted from each partition. The 2D co-occurrence matrix on every slice of the patch is then calculated from $R_{l_{p,i}}(p)$ and the following gray-level features are extracted: (1) autocorrelation, (2) contrast, (3) cluster shade, (4) dissimilarity, (5) energy, (6) entropy, (7) variance, (8) homogeneity, (9) correlation, (10) cluster prominence, and (11) inverse difference. To reduce the redundancy in the features, we use k-SVD dictionary learning \cite{aharon20064311}. A dictionary $D^p$ for every partition $p\in P$ is learned. Specifically, we begin by extracting the centerline of the initial ASM\textsubscript{p} segmentation using the shortest path graph. Afterwards, we choose the point $c_{p,i}$ on the centerline that is closest to the landmark $l_{p,i}$ in $l^2$-\textit{norm} sense. Subsequently, co-occurrence features are extracted from the patch $R_{c_{p,i}}(p)$. The likelihood of voxels belonging to the optic nerve is determined by using sparse representation classification (SRC) \cite{wright2009210}. In SRC framework, the classification problem is formulated as:
$\mathop {{\rm{argmin}}}\limits_\beta \left\| {f' - {D^p}\beta} \right\|_2^2 + \lambda {\left\| \beta \right\|_1}$, where $f'$ is the discriminative feature representation of the testing voxel, $\beta$ is the sparse code for the testing voxel, $\lambda$  is the coefficient of sparsity, and $r^p= f' - {D^p}{\beta^p}$ is the reconstruction residue of the sparse reconstruction. The likelihood $h$ of a testing voxel $y$ is calculated with the indicator function $h(\nu)$ with $h(\nu)=1$ if $r^p_y\le r^p_{y+1}$ and $-1$ otherwise, $r^p_{y}$ is the reconstruction residue at testing voxel $y$ and $r^p_{y+1}$ is the reconstruction residue at the neighboring next voxel to $y$ in the normal direction outwards from the centerline. To move landmark $l_{p,i}$ on the surface of the segmentation, we search in the normal direction. A position with the most similar profile pattern to the boundary pattern is chosen as the new position of the landmark using the following objective function,
$
\mathop {{\rm{arg max}}}\limits_h {\rm{ }}\left( {\mathop {{\rm{arg min}}}\limits_\delta  \left( {{{\left\| {P_{\left\{ { - 1,1} \right\}}^h\left( {{c_{p,i}} + \delta .\overrightarrow {{N_{{c_{p,i}}}}} } \right) - \overline P _{\left\{ { - 1,1} \right\}}^h} \right\|}_2}} \right){\rm{ + }}\frac{1}{|h|}{\rm{ }}} \right)| \delta  \in \left[ {0,A} \right],
$ where $\overline P _{\left\{ { - 1,1} \right\}}^h = \left[ {\underbrace { - 1, - 1,..., - 1}_{\left| h \right|},\underbrace {1,1,...,1}_{\left| h \right|}} \right]$ is the boundary pattern, $A$ is the search range, $N_{c_{p,i}}$ is outward normal direction at point $c_{p,i}$, $\delta$ is the position off-set to be optimized and $\overline P _{\left\{ { - 1,1} \right\}}^h$ is the desired boundary pattern. The length of the boundary pattern $|h|$ is desirable to be maximized to mitigate the effects of noise and false positives in the pattern.
  
\subsection{Hierarchical segmentation}
In order to enhance the robustness of the proposed method, we adopted a hierarchical segmentation approach by incorporating scale dependent information. The idea is that the coarser levels handles robustness while the finer-scale concentrates on the accuracy of the boundary. The segmentation at a coarser scale is subsequently used to initialize the finer scale. To achieve the hierarchical joint segmentation the following steps are adopted: (1) The number of shape partitions are dyadically increased from the coarsest to the finest scale. The number of partitions $n_j$ at the coarser scales $j$ are calculated as: $n_j=\lceil 2^{-j}G^J\rceil$, where $G^J$ is the number of partitions at the finest scale $J$. (2) The patch size used to calculate the appearance features (Section \ref{sec:fly}) are dyadically decreased from coarser to finer scales.

\section{Results}
\begin{figure}[!t]
\vspace{-0.2in}
\centering
\includegraphics[width=0.85\linewidth]{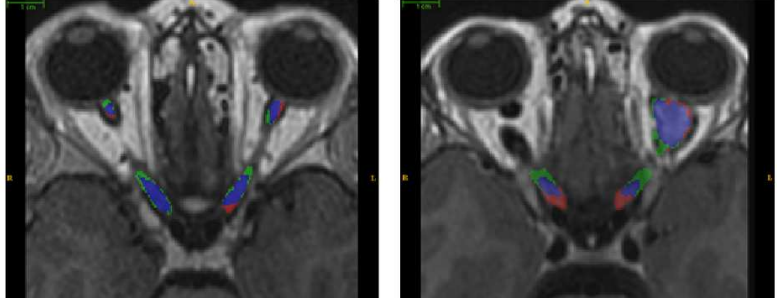}
\caption{\small{Segmentation results for a representative healthy (\emph{left}) and OPG case (\emph{right}). \emph{Blue} label shows overlap area of manual and automated segmentation, \emph{red} label shows the manual label while the \emph{green} label shows the automated segmentation.}}
\label{fig:results}
\vspace{-0.2in}
\end{figure}

After Institutional Review Board approval, 15 pediatric MRI scans with healthy AVPs and 6 with OPG were acquired for this study. The acquired data were T1 weighted cube with Gadolinium contrast enhancement having spatial resolution between $0.39\times 0.39\times 0.6\text{mm}^3$ to $0.47\times 0.47\times 0.6\text{mm}^3$. The manual ground truth for optic pathway segmentation was created by an expert neuro-radiologist and an expert neuro-ophthalmologist. During \textbf{the training stage}, the dataset was affinely registered to a randomly chosen reference image using a two-stage hierarchical approach: first by optimizing the registration parameters for the entire brain and later by optimizing over the region of interest around the optic nerve. The surfaces for each training instance were computed using the tetrahedral mesh generation approach followed by point set registration to the reference surface. Based on our training set, optimal number of partitions were found to be 12. Three hierarchical scales for shape model and appearance were used. The refinement model was learned \emph{on-the-fly} from the initial segmentation using a patch of size $11\times 11\times 11$ voxels at the coarsest level. The normalized derivative, the tissue intensity probability, and the tubular structure probability were used together as a unified feature set of size 33 to train the refinement model. To learn the sparse dictionary, co-occurrence features were extracted with an offset of 1 and four directions ($0$, $\frac{\pi}{4}$, $\frac{\pi}{2}$, $\frac{3\pi}{4}$). The co-occurrence features presented in Section \ref{sec:fly} are then calculated for each direction. During \textbf{the testing stage}, the test image was first registered to the randomly selected reference set followed by automatic overlapping partitioning. The mean shape of the training set was used to initialize the shape model. Fig. \ref{fig:results} shows the qualitative results of PAScAL against the ground truth manual segmentation.

\begin{figure*}[!b]
\vspace{-0.2in}
\centering
\includegraphics[width=0.43\linewidth]{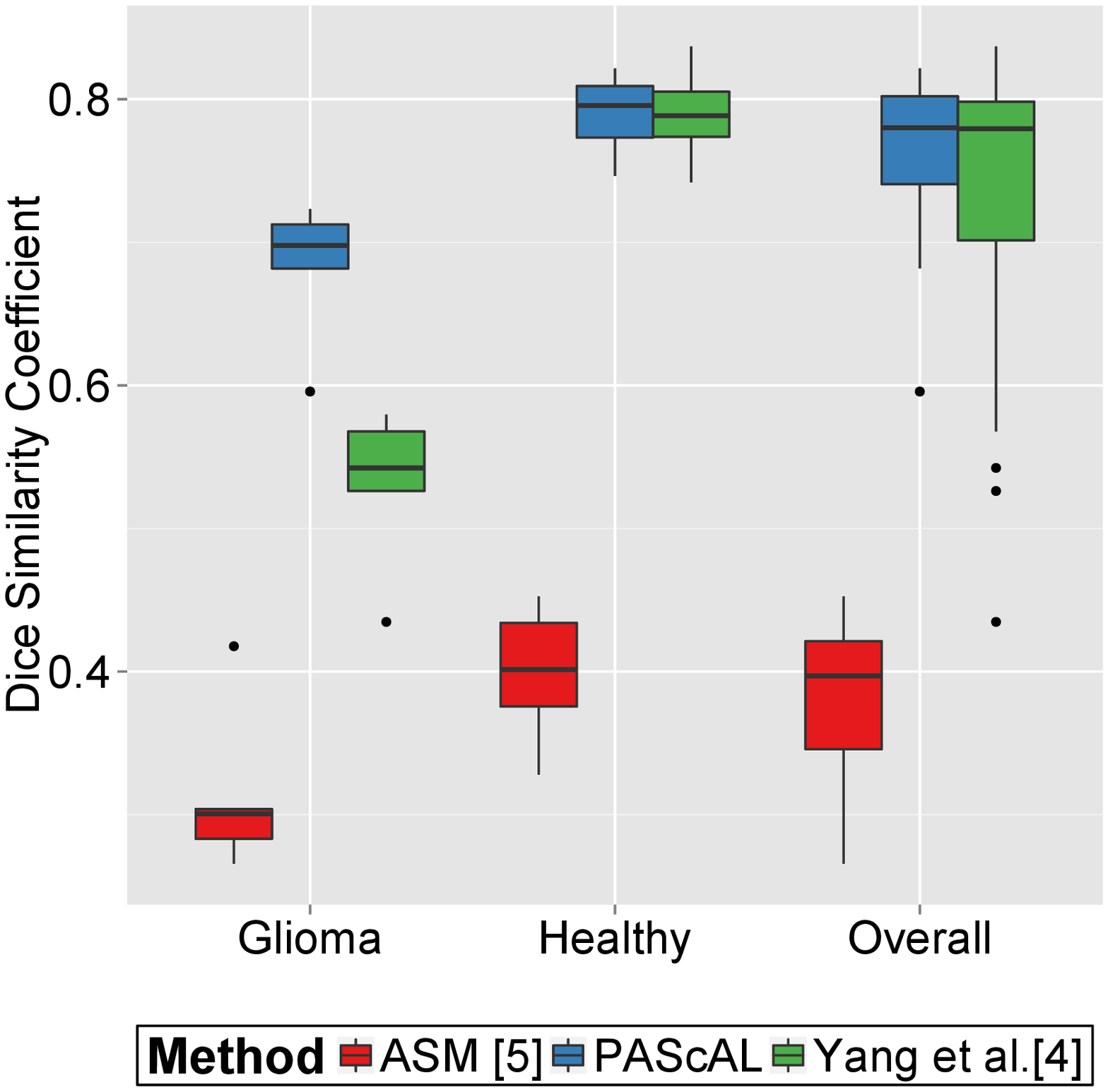}
\includegraphics[width=0.43\linewidth]{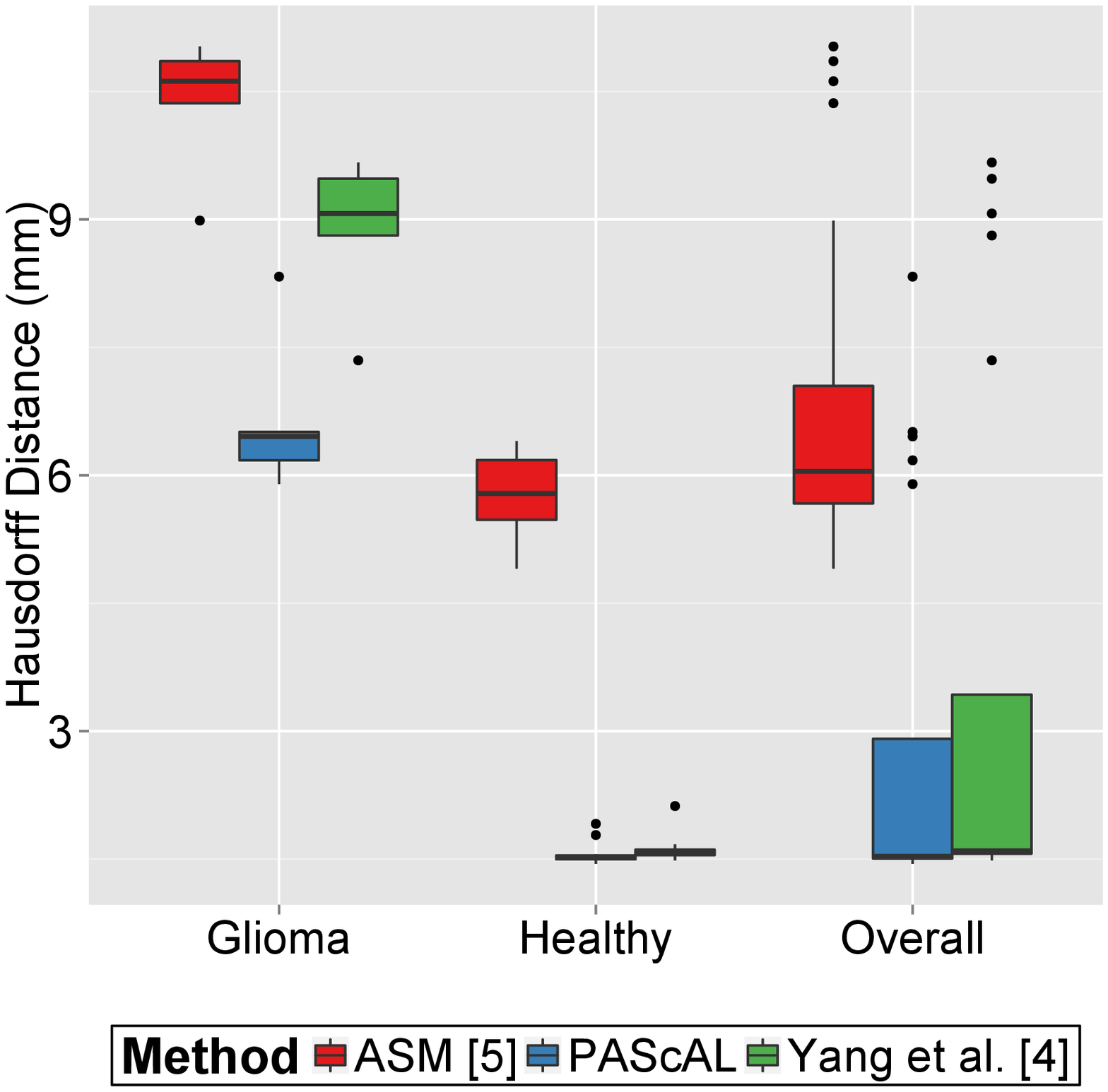}
\caption{\small{Quantitative comparison of PAScAL with traditional ASM and partitioned ASM method presented by Yang et al. \cite{Xue2014109}.}}
\label{fig:quantitative}
\vspace{-0.3in}
\end{figure*}

For quantitative evaluation, the Dice similarity coefficient (DSC) and Hausdorff distance (HD) were calculated between the segmentation obtained using PAScAL and the expert generated ground truth. The quantitative results based on the leave-one-out evaluation are reported in Fig. \ref{fig:quantitative}. An average DSC of $0.32$ for ASM, $0.53$ for Yang et al.'s approach \cite{Xue2014109}, and $0.68$ for PAScAL is obtained, showing significant improvement by PAScAL over both methods (p-value (\emph{Wilcoxon signed rank test}): ASM=$<0.001$, Yang's partitioned ASM=$0.015$).

\subsection{Automatic optic pathway glioma detection}
The demonstrated of the AVP is used to establish the clinical biomarker of the OPG based on the radius profile of the optic nerve. Specifically, the average radius of the optic nerve only (ref. Fig. \ref{fig:introduction} (c)) is calculated along the center-line of the training data set for healthy and OPG cases. A statistically significant difference between the average radii of the two classes was found based on the ground truth data (healthy optic nerve ($0.401\pm 0.050$mm), optic nerve with OPG ($0.800\pm 0.293$mm), p-value$<0.001$). No significant correlation between the average radius and the patient age, head circumference, and brain volume was found. To date, no established nomogram exist for the assessment of OPG; however, according to the World Health Organization osteopenia is diagnosed if the T score is $<1$ standard deviation ($\sigma$) from the mean of healthy population, osteoprosis is defined as $<2.5\sigma$ from the mean \cite{linguraru2012588}. Adopting similar approach, we define the detection of OPG in the optic nerve if the mean radius $>2.5\sigma$ from the mean of healthy population. Based on the adopted criteria, all 21 cases (15 healthy + 6 OPG cases) were classified with accuracy demonstrating the PAScAL to automatically detect pathologies of the optic nerve.      

\section{Conclusion}
{{We presented an automated technique, PAScAL, for the segmentation of anterior visual pathway from MRI scans of the brain based on partitioned shape models with sparse appearance learning. Our work addresses the challenge of segmenting cranial nerve pathways with shape and appearance variations due to unpredictable pathological changes.}} Experiments conducted using 21 T1 MRI scans, containing instances of both healthy and pathological cases, demonstrated superior performance of PAScAL over existing approaches. {{The application of PAScAL in segmenting anterior visual pathway shows its potential in analyzing other long and thin anatomical structures with pathologies.}} 

\bibliographystyle{splncs}
\bibliography{OpticNerveBib}
\end{document}